\documentclass[10pt,twocolumn,letterpaper]{article}

\usepackage{cvpr}
\usepackage{times}
\usepackage{epsfig}
\usepackage{graphicx}
\usepackage{amsmath}
\usepackage{amssymb}
\usepackage{fancyhdr}

\usepackage{enumitem} 

\DeclareMathOperator*{\argmin}{argmin}

\newcommand\blfootnote[1]{%
	\begingroup
	\renewcommand\thefootnote{}\footnote{#1}%
	\addtocounter{footnote}{-1}%
	\endgroup
}

\usepackage[pagebackref=true,breaklinks=true,letterpaper=true,colorlinks,bookmarks=false]{hyperref}
\usepackage{bm}
\cvprfinalcopy 

\def\cvprPaperID{695} 

\def\mathbi#1{\textbf{\em #1}}
\ifcvprfinal\fi
\begin{document}

\title{\vspace{-0.55cm} Recovering Realistic Texture in Image Super-resolution by \\ Deep Spatial Feature Transform}

\author{
	Xintao Wang $^{1}$ \hspace{9pt} Ke Yu$^{1}$ \hspace{9pt} Chao Dong$^{2}$ \hspace{9pt} Chen Change Loy$^{1}$\\
	\small{$^{1}$CUHK - SenseTime Joint Lab, The Chinese University of Hong Kong, 
	$^{2}$SenseTime Research}\\
	{\tt\small \{wx016, yk017, ccloy\}@ie.cuhk.edu.hk \hspace{5pt} dongchao@sensetime.com}\\
}

\maketitle
\pagestyle{fancy}
\fancyhead{}
\fancyfoot{}
\renewcommand{\headrulewidth}{0pt}
\cfoot{\thepage} 

\vspace{-0.7cm}
\begin{abstract}
Despite that convolutional neural networks (CNN) have recently demonstrated high-quality reconstruction for single-image super-resolution (SR), recovering natural and realistic texture remains a challenging problem.
In this paper, we show that it is possible to recover textures faithful to semantic classes.
In particular, we only need to modulate features of a few intermediate layers in a single network conditioned on semantic segmentation probability maps. This is made possible through a novel Spatial Feature Transform (SFT) layer that generates affine transformation parameters for spatial-wise feature modulation.
SFT layers can be trained end-to-end together with the SR network using the same loss function. During testing, it accepts an input image of arbitrary size and generates a high-resolution image with just a single forward pass conditioned on the categorical priors.
Our final results show that an SR network equipped with SFT can generate more realistic and visually pleasing textures in comparison to state-of-the-art SRGAN~\cite{ledig2017photo} and EnhanceNet~\cite{sajjadi2017enhancenet}. \blfootnote{All data, codes, models and supplementary material can be downloaded from \url{http://mmlab.ie.cuhk.edu.hk/projects/SFTGAN/}.}
\end{abstract}

\vspace{-0.4cm}
\section{Introduction}
\label{sec:introduction}

Single image super-resolution aims at recovering a high-resolution (HR) image from a single low-resolution (LR) one. 
The problem is ill-posed since a multiplicity of solutions exist for any given low-resolution pixel. 
To overcome this problem, contemporary methods such as those based on deep convolutional neural networks~\cite{dong2014learning,dong2016image,dong2016accelerating,kim2016accurate,lai2017deep,ledig2017photo,kim2016deeply,tai2017image,tai2017memnet} constrain the solution space through learning mapping functions from external low- and high-resolution exemplar pairs.
To push the solution closer to the natural manifold, new losses are proposed to replace the conventional pixel-wise mean squared error (MSE) loss~\cite{dong2014learning} that tends to encourage blurry and overly-smoothed results.
Specifically, perceptual loss~\cite{johnson2016perceptual, bruna2015super} is introduced to optimize a super-resolution model in a feature space instead of pixel space. Ledig \etal~\cite{ledig2017photo} and Sajjadi \etal~\cite{sajjadi2017enhancenet} further propose adversarial loss to encourage the network to favor solutions that look more like natural images. With these loss functions the overall visual quality of reconstruction is significantly improved.

 \begin{figure}[t]
    \centering
    \includegraphics[width=\linewidth]{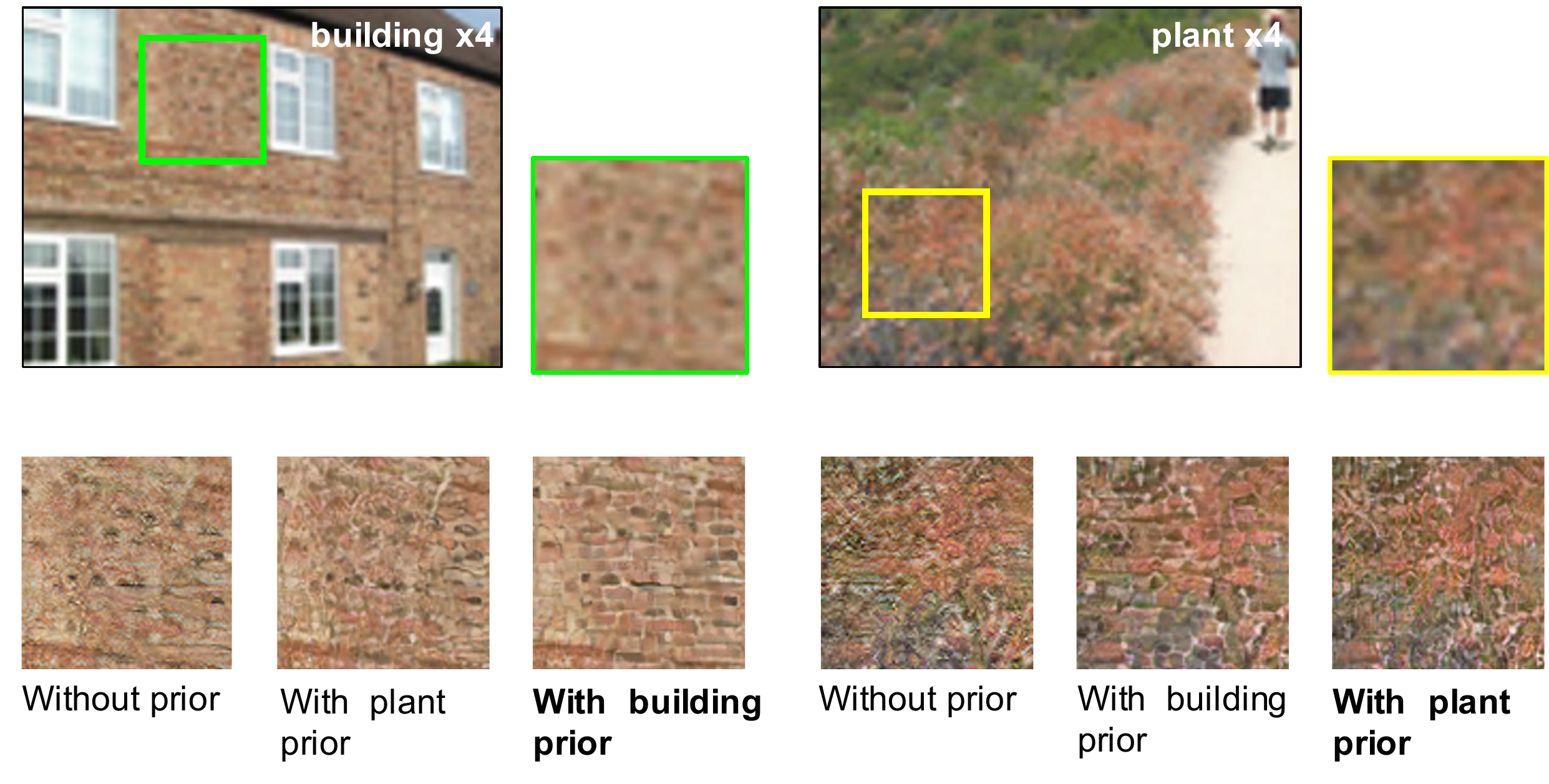}
    \caption{\small
        The extracted building and plant patches from two low-resolution images look very similar. Using adversarial loss and perceptual loss without prior could add details that are not faithful to the underlying class. More realistic results can be obtained with the correct categorical prior. (\textbf{Zoom in for best view}). }
    \label{fig:building_plant_semantic_matters}
    \vspace{-0.4cm}
\end{figure}

Though great strides have been made, texture recovery in SR remains an open problem. Examples are shown in Fig.~\ref{fig:building_plant_semantic_matters}. 
A variety of different HR patches could have very similar LR counterparts, as shown by the building and plant examples.
Generating realistic textures faithful to the inherent class is non-trivial. 
The results obtained by using perceptual and adversarial losses (without prior) do add fine details to the reconstructed HR image. But if we examine closely, these details are not reminiscent of the textures one would usually observe. 
Without stronger prior information, existing methods struggle in distinguishing these LR patches and restoring natural and realistic textures thereon.

\begin{figure*}[t]
\begin{center}
	\includegraphics[width=\linewidth]{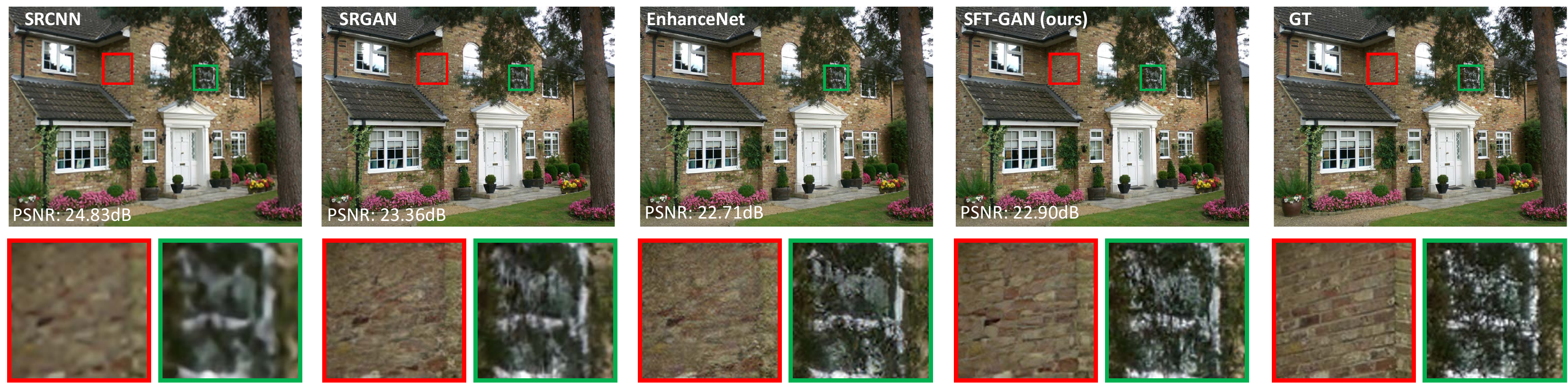}
	\caption{\small
		Comparing different SR approaches with downsampling factor $\times 4$: SRCNN~\cite{dong2014learning}, SRGAN~\cite{ledig2017photo}, EnhanceNet~\cite{sajjadi2017enhancenet}, our proposed SFT-GAN and the original HR image. SRGAN, EnhanceNet, and SFT-GAN clearly outperform SRCNN in terms of perceptual quality, although they yield lower peak signal-to-noise ratio (PSNR) values. SRGAN and EnhanceNet result in more monotonous textures across different patches while SFT-GAN is capable of generating richer and visually pleasing textures. (\textbf{Zoom in for best view}).}
	\label{fig:intro_examples}
	\vspace{-0.7cm}
\end{center}
\end{figure*}

We believe that the categorical prior, which characterizes the semantic class of a region in an image (\eg,~sky, building, plant), is crucial for constraining the plausible solution space in SR.
We demonstrate the effectiveness of categorical prior using the same example in Fig.~\ref{fig:building_plant_semantic_matters}.
Specifically, we try to restore the visually ambiguous plant and building pairs using two different CNN models, each of which is specially trained on a plant dataset and a building dataset. 
It is observed that generating realistic textures faithful to the inherent class can be better achieved by selecting the correct class-dedicated model.
This phenomenon is previously documented by Timofte \etal~\cite{timofte2016semantic}. They train specialized models for each semantic category on exemplar-based methods~\cite{zeyde2010single, timofte2014a+} and show that SR results can be improved by semantic priors.

In this study, we wish to investigate class-conditional image super-resolution with CNN. This problem is challenging especially when multiple segments of different classes and sizes co-exist in a single image. No previous work has investigated how categorical priors can be obtained and further incorporated into the reconstruction process.
We make this attempt by exploring the possibility of using semantic segmentation maps as the categorical prior. Our experiments embrace this choice and show that segmentation maps encapsulate rich categorical prior up to pixel level. In addition, semantic segmentation results on LR images are satisfactory given a contemporary CNN~\cite{long2015fully, liu2017deep,li2017not} that is fine-tuned on LR images.
The remaining question is to find a formulation that allows factorized texture generation in an SR network conditioned on segmentation maps.
This is a non-trivial problem. Training a separate SR model for each semantic class is neither scalable nor computationally efficient. Combining LR images with segmentation maps as inputs, or concatenating segmentation maps with intermediate deep features cannot make an effective use of segmentation.
%

In this work, we present a novel approach known as Spatial Feature Transform (SFT) that is capable of altering the behavior of an SR network through just transforming the features of some intermediate layers of the network.
Specifically, an SFT layer is conditioned on semantic segmentation probability maps, based on which it generates a pair of modulation parameters to apply affine transformation spatially on feature maps of the network.
The advantages of SFT are three-fold:
(1) It is parameter-efficient. Reconstruction of an HR image with rich semantic regions can be achieved with just a single forward pass through transforming the intermediate features of a single network. 
(2) SFT layers can be easily introduced to existing SR network structures. The layers can be trained end-to-end together with the SR network using conventional loss functions.
(3) It is extensible. While we consider categorical prior in our study, other priors such as depth maps can also be applied using the proposed SFT layer.
We demonstrate the effectiveness of our approach, named as SFT-GAN, in Fig.~\ref{fig:intro_examples}. 
More results, a user study, and ablation experiments are provided in Sec.~\ref{sec:experiments}.

\section{Related Work}
\label{sec:related_work}

\noindent \textbf{Single image super-resolution}.
Many studies have introduced prior information to help address the ill-posed SR problem.
Early methods explore a smoothing prior such as bicubic interpolation and Lanczos resampling~\cite{duchon1979lanczos}. 
Image priors such as edge features~\cite{fattal2007image, sun2008image}, statistics~\cite{kim2010single, aly2005image} and internal patch recurrence~\cite{glasner2009super} are employed to improve performance. 
Dong \etal~\cite{dong2011image} train domain specific dictionaries to better recover local structures in a sparse representation framework. 
Sun \etal~\cite{sun2010context} propose context-constrained super-resolution by learning from
texturally similar training segments. Timofte \etal~\cite{timofte2016semantic} investigate semantic priors by training specialized models separately for each semantic category on exemplar-based methods~\cite{zeyde2010single, timofte2014a+}. 
In contrast to these studies, we explore categorical priors in the form of segmentation probability maps in a CNN framework.

Contemporary SR algorithms are mostly learning-based methods, including neighbor embedding~\cite{chang2004super}, sparse coding~\cite{yang2008image, zeyde2010single, timofte2013anchored, timofte2014a+} and random forest~\cite{schulter2015fast}. 
As an instantiation of learning-based methods, Dong \etal~\cite{dong2014learning} propose SRCNN for learning the mapping of LR and HR images in an end-to-end manner. Later on, the field has witnessed a variety of network architectures, such as a deeper network with residual learning~\cite{kim2016accurate}, Laplacian pyramid structure~\cite{lai2017deep}, residual blocks~\cite{ledig2017photo}, recursive learning~\cite{kim2016deeply,tai2017image}, and densely connected network~\cite{tai2017memnet}. Multi-scale guidance structure has also been proposed for depth map super-resolution~\cite{hui2016depth}.
Different losses have also been proposed. Pixel-wise loss functions, like MSE and Charbonnier penalty~\cite{lai2017deep}, encourage the network to find an average of many plausible solutions and lead to overly-smooth results. Perceptual losses~\cite{johnson2016perceptual, bruna2015super} are proposed to enhance the visual quality by minimizing the error in a feature space. Ledig \etal~\cite{ledig2017photo} introduce an adversarial loss, generating images with more natural details. Sajjadi~\etal~\cite{sajjadi2017enhancenet} develop a similar approach and further explore the local texture matching loss, partly reducing visually unpleasant artifacts. 
We use the same losses but encourage the network to find solutions under the categorical priors. 
Enforcing category-specific priors in CNN has been attempted in Xu~\etal~\cite{xu2017learning} but they only focus on two classes of images, \ie, faces and text. Prior is assumed at image-level rather than pixel-level. We take a further step to assume multiple categorical classes to co-exist in an image, and propose an effective layer that enables an SR network to generate rich and realistic textures in a single forward pass conditioned on the prior provided up to the pixel level.

\begin{figure*}[t]
	\centering
	\includegraphics[width=1\linewidth]{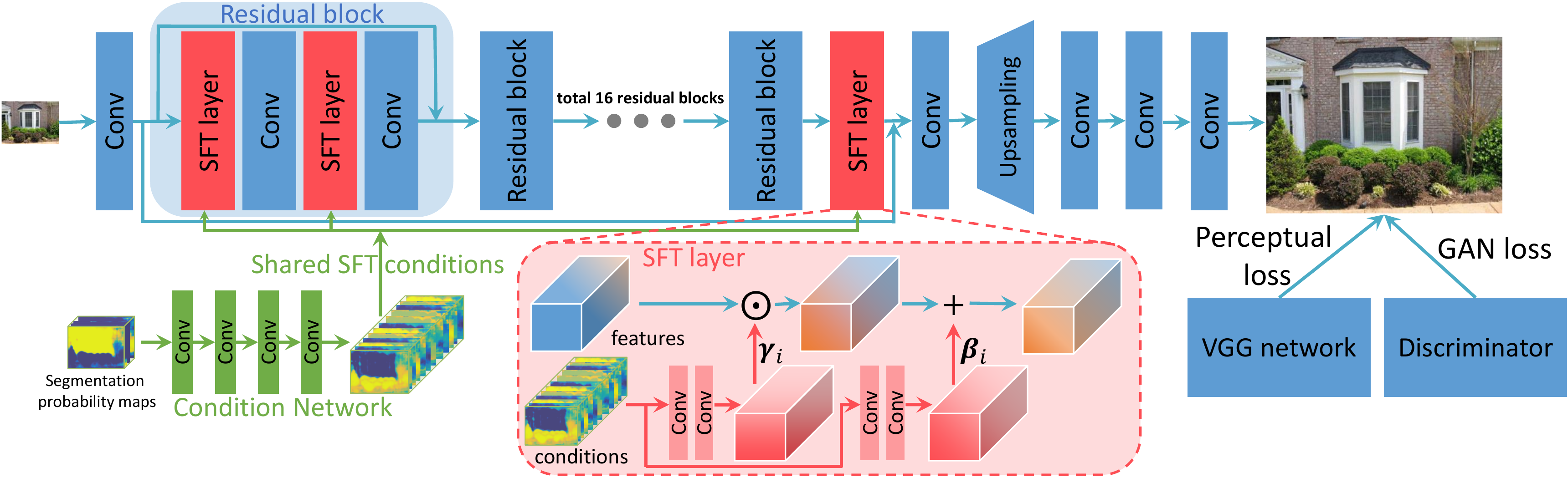}
	\caption{\small
		The proposed SFT layers can be conveniently applied to existing SR networks. All SFT layers share a condition network. The role of the condition network is to generate intermediate conditions from the prior, and broadcast the conditions to all SFT layers for further generation of modulation parameters.
	}
	\label{fig:framework}
	\vspace{-0.3cm}
\end{figure*}

\noindent \textbf{Network conditioning}.
Our work is inspired by previous studies on feature normalization.
Batch normalization (BN) is a widely used technique to ease network training by normalizing feature statistics~\cite{ioffe2015batch}. 
Conditional Normalization (CN) applies a learned function of some conditions to replace parameters for feature-wise affine transformation in BN. 
Some variants of CN have proven highly effective in image style transfer~\cite{dumoulin2016learned, huang2017arbitrary, ghiasi2017exploring}, visual question answering~\cite{de2017modulating} and visual reasoning~\cite{perez2017learning}.
Perez \etal~\cite{perez2017film} develop a feature-wise linear modulation layer (FiLM),  to exploit linguistic information for visual reasoning.
This layer can be viewed as a generalization of CN. Perez \etal show that the affine transformation in CN needs not be placed after normalization. Features can be directly modulated.
FiLM shows promising results in visual reasoning. Nonetheless, the formulation cannot handle conditions with spatial information (\eg,~semantic segmentation maps) since FiLM accepts a single linguistic input and outputs one scaling and one shifting parameter for each feature map, agnostic to spatial location. 
Preserving spatial information is crucial for low-level tasks, \eg, SR, since these tasks usually require adaptive processing at different spatial locations of an image. Applying FiLM in SR will result in homogeneous spatial feature modulation, hurting SR quality. 
The proposed SFT layer addresses this shortcoming. It is capable of converting spatial conditions for not only feature-wise manipulation but also spatial-wise transformation. 

\noindent \textbf{Semantic guidance}.
%
In image generation~\cite{isola2017image, chen2017photographic}, semantic segments are used as input conditions to generate natural images. Gatys \etal~\cite{gatys2017controlling} use semantic maps to control perceptual factors in neural style transfer. 
\cite{ren2017video} uses semantic segmentation for video deblurring.
Zhu \etal~\cite{zhu2017your} propose an approach to generate new clothing on a wearer. It first generates human segmentation maps and then uses them to render plausible textures by enforcing region-specific texture rendering. Our work differs from these works mainly in two aspects. First, we use semantic maps to guide texture recovery for different regions in SR domain. Second, we utilize probability maps to capture delicate texture distinction instead of simple image segments. 

\section{Methodology}
\label{sec:methodology}

Given a single low-resolution image $\bm{x}$, super-resolution aims at estimating a high-resolution one $\hat{\bm{y}}$, which is as similar as possible to the ground truth image $\bm{y}$. 
Current CNN-based methods use feed-forward networks to directly learn a mapping function $G_{\bm{\theta}}$ parametrized by $\bm{\theta}$ as
\begin{equation}\label{equ:y_hat}
\hat{\bm{y}} = G_{\bm{\theta}}(\bm{x}).
\end{equation}
In order to estimate $\hat{\bm{y}}$, a specific loss function $\mathcal{L}$ is designed for SR to optimize $G_{\bm{\theta}}$ on the training samples, 
\begin{equation}\label{equ:loss_function}
\hat{\bm{\theta}} = \argmin_{\bm{\theta}} \sum\nolimits_{i} \mathcal{L} (\hat{\bm{y}_{i}}, \bm{y}_{i}),
\end{equation}
where $(\bm{x}_i, \bm{y}_i)$ are training pairs. 
Perceptual loss~\cite{johnson2016perceptual, bruna2015super} and adversarial loss~\cite{ledig2017photo, sajjadi2017enhancenet} are introduced to solve the regression-to-the-mean problem that is usually caused by conventional MSE-oriented loss functions. These new losses greatly improve the perceptual quality of reconstructed images. However, the generated textures tend to be monotonous and unnatural, as seen in Fig.~\ref{fig:building_plant_semantic_matters}.

We argue that the semantic categorical prior, \ie,~knowing which region belongs to the sky, water, or grass, is beneficial for generating richer and more realistic textures. The categorical prior $\Psi$ can be conveniently represented by semantic segmentation probability maps $\mathbf{P}$,
\begin{equation}\label{equ:probability_maps}
\Psi = \mathbf{P} = (P_1, P_2,\ldots,P_k,\ldots,P_K),
\end{equation}
where $P_k$ is the probability map of $k^{th}$ category and $K$ is the total number of considered categories.
To introduce priors in SR, we reformulate Eqn.~\eqref{equ:y_hat} as
\begin{equation}\label{equ:general_prior}
\hat{\bm{y}} = G_{\bm{\theta}}(\bm{x}|\Psi),
\end{equation}
where $\Psi$ defines the prior upon which the mapping function can condition.
Note that apart from categorical priors, the proposed formulation is also applicable to other priors such as depth information, which could be helpful to the recovery of texture granularity in SR. And one could easily extend the formulation to consider multiple priors simultaneously.
In the following section, we focus on categorical priors and the way we use them to influence the behavior of an SR network.

\subsection{Spatial Feature Transform}
\label{subsec:sfm}

A Spatial Feature Transform (SFT) layer learns a mapping function $\mathcal{M}$ that outputs a modulation parameter pair $(\bm{\gamma}, \bm{\beta})$ based on some prior condition $\Psi$. 
The learned parameter pair adaptively influences the outputs by applying an affine transformation spatially to each intermediate feature maps in an SR network. During testing, only a single forward pass is needed to generate the HR image given the LR input and segmentation probability maps.
  
More precisely, the prior $\Psi$ is modeled by a pair of affine transformation parameters $(\bm{\gamma}, \bm{\beta})$ through a mapping function $\mathcal{M}: \Psi \mapsto (\bm{\gamma}, \bm{\beta})$.
Consequently,
\begin{equation}\label{equ:general_prior_with_parameters}
\hat{\bm{y}} = G_{\bm{\theta}}(\bm{x}|\bm{\gamma}, \bm{\beta}), \quad  (\bm{\gamma}, \bm{\beta}) = \mathcal{M}(\Psi).
\end{equation}
After obtaining $(\bm{\gamma}, \bm{\beta})$ from conditions, the transformation is carried out by scaling and shifting feature maps of a specific layer:
\begin{equation}\label{eqn:SFT_layer}
\text{SFT}(\mathbi{F}|\bm{\gamma}, \bm{\beta}) = \bm{\gamma} \odot \mathbi{F} + \bm{\beta},
\end{equation}
where $\mathbi{F}$ denotes the feature maps, whose dimension is the same as $\bm{\gamma}$ and $ \bm{\beta}$, and $\odot$ is referred to element-wise multiplication, \ie, Hadamard product.
Since the spatial dimensions are preserved, the SFT layer not only performs feature-wise manipulation but also spatial-wise transformation.

Figure~\ref{fig:framework} shows an example of implementing SFT layers in an SR network. We provide more details of the SR branch in Sec.~\ref{subsec:architecture}. Here we focus on the conditioning part.
The mapping function $\mathcal{M}$ can be arbitrary functions. In this study, we use a neural network for $\mathcal{M}$ so that it can be optimized end-to-end with the SR branch. To further share parameters among multiple SFT layers for efficiency, we use a small condition network to generate shared intermediate conditions that can be broadcasted to all the SFT layers. Meanwhile, we still keep few parameters inside each SFT layer to further adapt the shared conditions to the specific parameters $\bm{\gamma}$ and $\bm{\beta}$, providing fine-grained control to the features.

\noindent
\textbf{Segmentation probability maps as prior}.
We provide a brief discussion on the segmentation network we used. The details are provided in the \textit{supplementary material}. The LR image is first upsampled to the desired HR size with bicubic interpolation. It is then fed into a segmentation network~\cite{liu2017deep} as the input. The network is pretrained on the COCO dataset~\cite{lin2014microsoft} and then fine-tuned on the ADE dataset~\cite{zhou2017scene} with additional animal and mountain images. We train the network separately from the main SR network.

\begin{figure}[t]
    \begin{center}
       \includegraphics[width=1\linewidth]{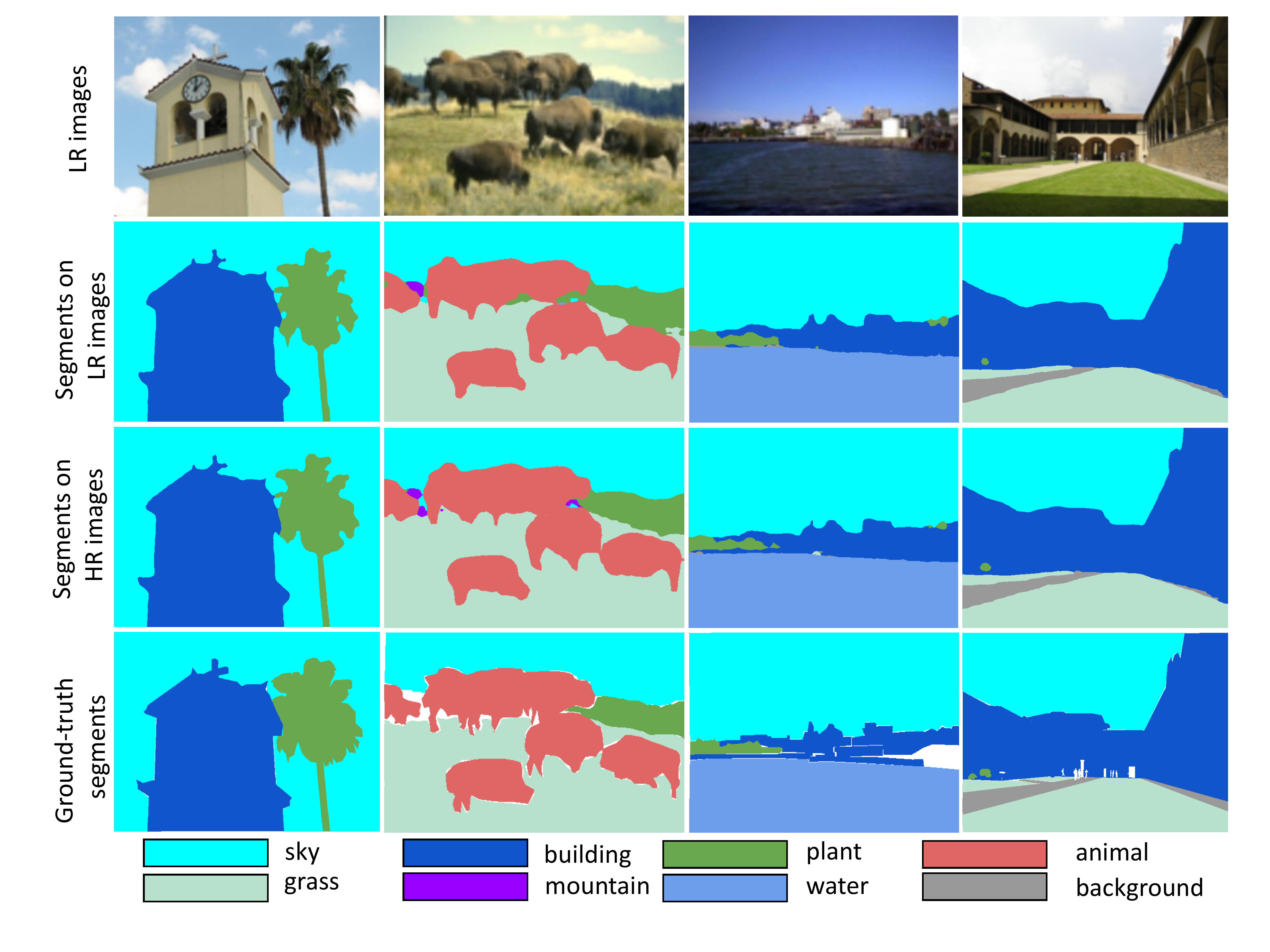}
    \end{center}
	\vspace{-0.25cm}
    \caption{Some examples on segmentation. First row: LR images. Second row: segmentation results on LR images. Third row: segmentation results on HR images. Forth row: Ground-truth segmentation.}
    \label{fig:segmentation}
    \vspace{-0.3cm}
\end{figure}

For a sanity check, we study the accuracy of segmentation maps obtained from LR image. In a typical setting of SR studies, LR images are downsampled with a scaling factor of $\times 4$ from HR images. We find that under this resolution, satisfactory segmentation results can still be obtained even on LR images given a modern CNN-based segmentation model~\cite{long2015fully, liu2017deep}.  Some LR images and the corresponding segmentation results are depicted in Fig.~\ref{fig:segmentation}. 
As can be observed in Fig.~\ref{fig:segmentation}, LR segmentation is close to that of HR. We have not yet tried segmentation on small objects as this remains a challenging problem in the image segmentation community. 
During testing, classes that fall outside the pre-defined $K$ segmentation classes will be categorized as `background' class. In this case, our method would still generate a set of default $\bm{\gamma}$ and $\bm{\beta}$, degenerating itself as SRGAN, \ie, treating all classes equally.

\noindent
\textbf{Discussion}.
There are alternative ways to introduce categorical priors to an SR network. For instance, one can concatenate the segmentation probability maps with the input LR image as a joint input to the network. We find that this method is ineffective in altering the behavior of CNN (see Sec.~\ref{subsec:ablation}). Another method is to directly concatenate the probability maps with feature maps in the SR branch. This approach resembles the multi-texture synthesis network~\cite{li2017diversified}\footnote{The approach in~\cite{li2017diversified} is not applicable in our work since it can only generate outputs with a fixed size due to the upsampling operation from one-hot vector.}. 
This method, though not as parameter efficient as SFT, amounts to simply adding a post-layer for feature-wise conditional bias. It is thus a special case of SFT.
Another more brute-force approach is to first decompose the LR image based on the predicted semantic class and process each region separately using a model trained specifically for that class.  These models may share features to save computation. The final output is generated by combining the output of each model class-wise. 
This method is computationally inefficient as we need to perform forward passes with several CNN models for a single input image.

\subsection{Architecture}
\label{subsec:architecture}
Our framework is based on adversarial learning, inspired by~\cite{ledig2017photo, sajjadi2017enhancenet}. Specifically, it consists of one generator $G_{\bm{\theta}}$ and one discriminator $D_{\bm{\eta}}$, parametrized by $\bm{\theta}$ and $\bm{\eta}$ respectively. They are jointly trained with a learning objective given below:
\begin{equation*}\label{equ:gan_optimization}
\min_{\bm{\theta}} \max_{\bm{\eta}} \mathbb{E}_{\bm{y}\sim p_{\text{HR}}}\log D_{\bm{\eta}}(\bm{y}) + \mathbb{E}_{\bm{x}\sim p_{\text{LR}}}\log (1-D_{\bm{\eta}}(G_{\bm{\theta}}(\bm{x}))),
\end{equation*}
where $p_{\text{HR}}$ and $p_{\text{LR}}$ are the empirical distributions of HR and LR training samples, respectively. 

The architecture of $G_{\bm{\theta}}$ is shown in Fig.~\ref{fig:framework}. 
It consists of two streams: a condition network and an SR network. 
The condition network takes segmentation probability maps as input, which are then processed by four convolution layers. It generates intermediate conditions shared by all the SFT layers. To avoid interference of different categorical regions in one image, we restrict the receptive field of the condition network by using $1\times1$ kernels for all the convolution layers.

The SR network is built with 16 residual blocks with the proposed SFT layers, which take the shared conditions as input and learn $(\bm{\gamma}, \bm{\beta})$ to modulate the feature maps by applying affine transformation. Skip connection~\cite{ledig2017photo} is used to ease the training of deep CNN. We upsample features by using nearest-neighbor upsampling followed by a convolution layer. The upsampling operation is performed in the latter part of the network and thus most computation is done in the LR space. 
Although we have not tried other architectures for the SR network, we believe many contemporary models such as DRRN~\cite{tai2017image} and MemNet~\cite{tai2017memnet} are applicable and can equally be benefited from the SFT layer.

For discriminator $D_{\bm{\eta}}$, we apply a VGG-style~\cite{simonyan2014very} network of strided convolutions to gradually decrease the spatial dimensions. The full architecture and details are provided in the \textit{supplementary material}.

\begin{figure*}[t]
	\begin{center}
		\includegraphics[width=1\linewidth]{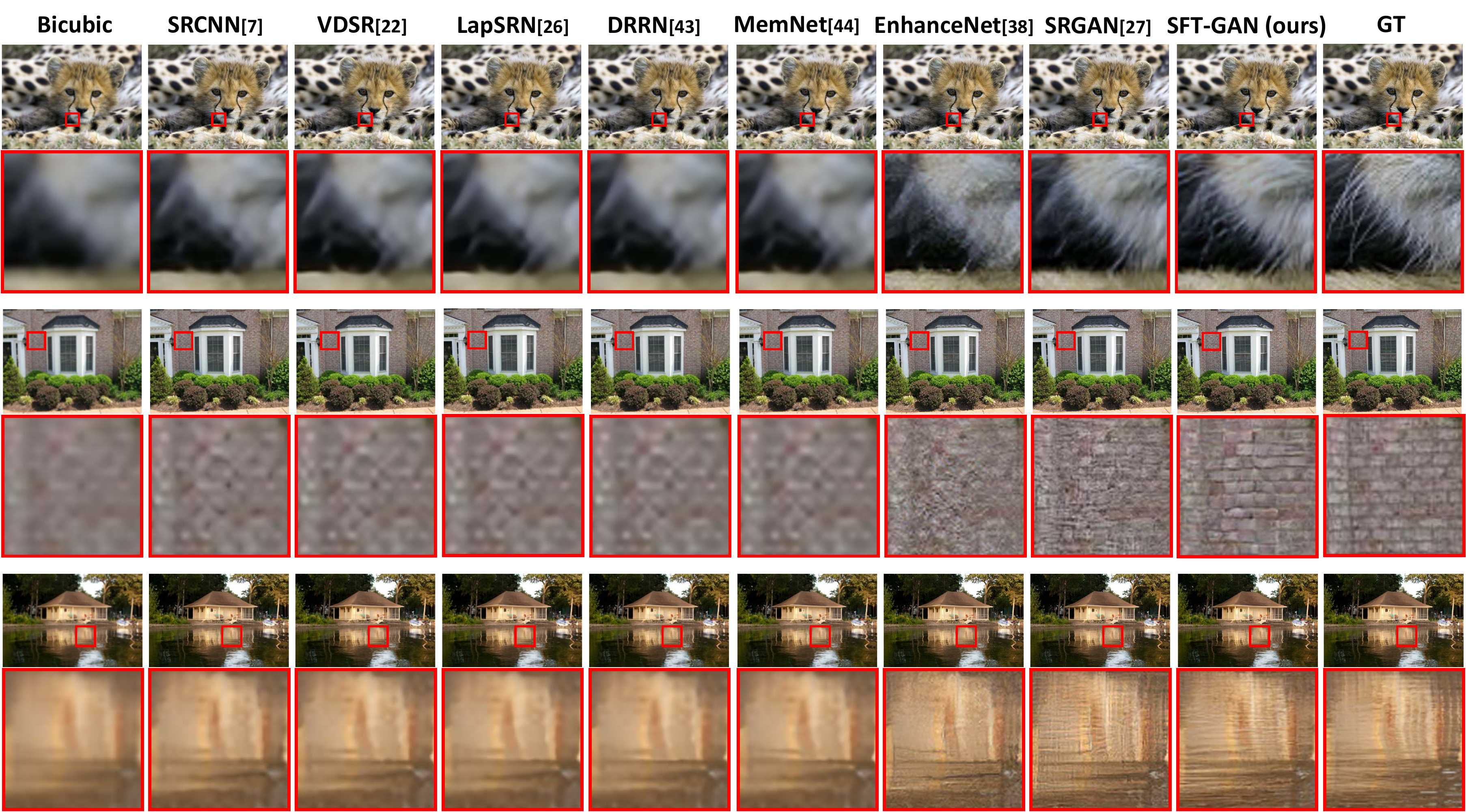}
		\caption{GAN-based methods (SRGAN~\cite{ledig2017photo}, EnhanceNet~\cite{sajjadi2017enhancenet} and ours) clearly outperform PSNR-oriented approaches in term of perceptual quality. Our proposed SFT-GAN is capable of generating richer and more realistic textures among different categories. The first and second restored images show that our method captures the characteristics of animal fur and building brick. In the third image, SRGAN tends to produce unpleasant water waves. (\textbf{Zoom in for best view}).}
		\label{fig:qualitative_examples}
		\vspace{-0.5cm}
	\end{center}
\end{figure*}

\subsection{Loss Function}
We draw inspiration from~\cite{ledig2017photo, sajjadi2017enhancenet} and apply perceptual loss and adversarial loss in our model. The perceptual loss measures the distance in a feature space. To obtain the feature maps, we use a pre-trained 19-layer VGG network~\cite{simonyan2014very}, denoted as $\phi$,
\begin{equation}\label{equ:perceptual_loss}
\mathcal{L}_P = \sum\nolimits_{i} \|\phi(\hat{\bm{y}_{i}})-\phi(\bm{y}_{i}) \|_2^2.
\end{equation}
Similar to~\cite{ledig2017photo}, we use the feature maps obtained by the forth convolution before the fifth max-pooling layer and compute the MSE on their feature activations.


The adversarial loss $\mathcal{L}_D$ from GAN is also used to encourage the generator to favor the solutions in the manifold of natural images, 
\begin{equation}\label{equ:adversarial_loss}
\mathcal{L}_D = \sum\nolimits_{i} \log (1-D_{\bm{\eta}}(G_{\bm{\theta}}(\bm{x}_{i}))).
\end{equation}

\section{Experiments}
\label{sec:experiments}

\noindent
\textbf{Implementation details}.
In this work, we focus on outdoor scenes since their textures are rich and well-suited for our study. For example, the sky is smooth and lacks sharp edges, while the building is rich of geometric patterns. The water presents smooth surface with waves, while the grass has matted textures. We assume seven categories, \ie, sky, mountain, plant, grass, water, animal and building. A `background' category is used to encompass regions that do not appear in the aforementioned categories. 

Following~\cite{ledig2017photo}, all experiments were performed with a scaling factor of $\times 4$ between LR and HR images.
We initialized the SR network by parameters pre-trained with perceptual loss and GAN loss on ImageNet dataset. After removing low resolution images of size below 30kB, we obtained roughly 450k training images.
During training, we followed existing studies~\cite{dong2014learning, ledig2017photo} to obtain LR images by downsampling HR images using MATLAB bicubic kernel. 
The mini-batch size was set to 16. The spatial size of cropped HR and LR sub-images were $96\times96$ and $24\times24$, respectively.

After initialization, with the same training setting, we fine-tuned our full network on outdoor scenes conditionally on the input segmentation probability maps.
In particular, we collected a new outdoor dataset by querying images from search engines using the defined categories as keywords. The dataset was divided into training and test partitions (OutdoorSceneTrain and OutdoorSceneTest)
For OutdoorSceneTrain, we cropped each image so that only one category exists, resulting in 1k to 2k images for each category. Background images were randomly sampled from ImageNet. The total number of training images is 10,324. The OutdoorSceneTest partition consists of 300 images and they are not pre-processed.
Segmentation probability maps were generated by the segmentation network (Sec.~\ref{subsec:sfm}). 

For optimization, we used Adam~\cite{kingma2014adam} with $\beta_1=0.9$. The learning rate was set to $1\times 10^{-4}$ and then decayed by a factor of 2 every 100k iterations. We alternately optimized the generator and discriminator until the model converged at about $5\times 10^{5}$ iterations.
Inspired by~\cite{odena2016conditional}, our discriminator not only distinguishes whether the input is real or fake, but also predicts which category the input belongs to. This is possible since our training images were cropped to contain only one category. This restriction was not applied on test images. We find this strategy facilitates the generation of images with more realistic textures.

We did not conduct our main evaluation on standard benchmarks such as Set5~\cite{bevilacqua2012low}, Set14~\cite{zeyde2010single} and BSD100~\cite{martin2001database} since these datasets are lack of regions with well-defined categories. Nevertheless, we will show that our method still perform satisfactorily on out-of-category examples in Sec.~\ref{subsec:ablation}. Results (PSNR, SSIM) on standard benchmarks are provided in the \textit{supplementary material}.

\subsection{Qualitative Evaluation}

Figure~\ref{fig:qualitative_examples} shows the qualitative results of different models including PSNR-oriented methods, such as SRCNN~\cite{dong2014learning}, VDSR~\cite{kim2016accurate}, LapSRN~\cite{lai2017deep}, DRRN~\cite{tai2017image}, MemNet~\cite{tai2017memnet}, and GAN-based methods, such as SRGAN~\cite{ledig2017photo} and EnhanceNet~\cite{sajjadi2017enhancenet}. More results are provided in the \textit{supplementary material}. For SRGAN, we re-implemented their method and fine-tuned the model with the same setting as ours. We directly used the released test code of EnhanceNet since no training code is available. Despite of preserving sharp edges, PSNR-oriented methods always produce blurry textures. SRGAN and EnhanceNet largely improve the high-frequency details, however, they tend to generate monotonous and unnatural textures, like water waves in Fig.~\ref{fig:qualitative_examples}. Our method employs categorical priors to help capture the characteristics of each category, leading to more natural and realistic textures.

\subsection{User Study}
\label{subsec:user_study}

We performed a user study to quantify the ability of different approaches to reconstruct perceptually convincing images. To better compare our method against PSNR-oriented baselines and GAN-based approaches, we divided the evaluations into two sessions.
In the first session, we focused on PSNR-oriented baselines. The users were requested to rank 4 versions of each image: SRCNN~\cite{dong2014learning}, MemNet~\cite{tai2017memnet} (the state-of-the-art PSNR-oriented method), our SFT-GAN, and the original HR image according to their visual quality. We used 30 random images chosen from OutdoorSceneTest and all images were presented in a randomized fashion.
In the second session, we focused on GAN-based methods so that the user can concentrate on the texture quality. The subjects were shown the super-resolved image pairs (enlarged texture patches were depicted to facilitate the comparison). Each pair consists of an image of the proposed SFT-GAN and the counterpart generated by SRGAN~\cite{ledig2017photo} or EnhanceNet~\cite{sajjadi2017enhancenet}. The users were asked to pick the image with more natural and realistic textures. This session involved 96 randomly selected images in total.

\begin{figure}[t]
	\begin{center}
		\includegraphics[width=0.9\linewidth]{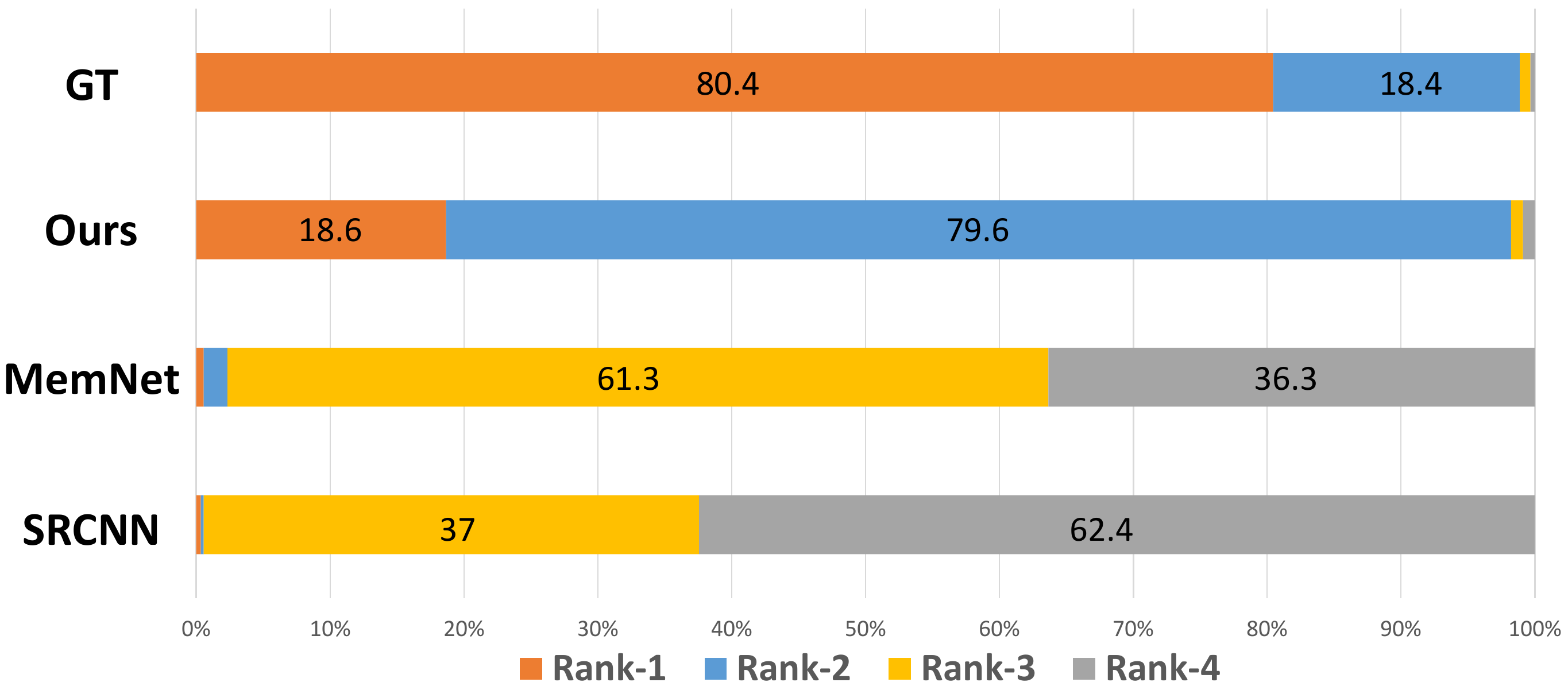}
		\caption{User study results of ranking SRCNN~\cite{dong2014learning}, MemNet~\cite{tai2017memnet}, SFT-GAN (ours), and the original HR image. Our method outperforms PSNR-oriented methods by a large margin. }
		\label{fig:user_study_rank}
		\vspace{-0.8cm}
	\end{center}
\end{figure}

\begin{figure}[t]
	\begin{center}
		\includegraphics[width=0.9\linewidth]{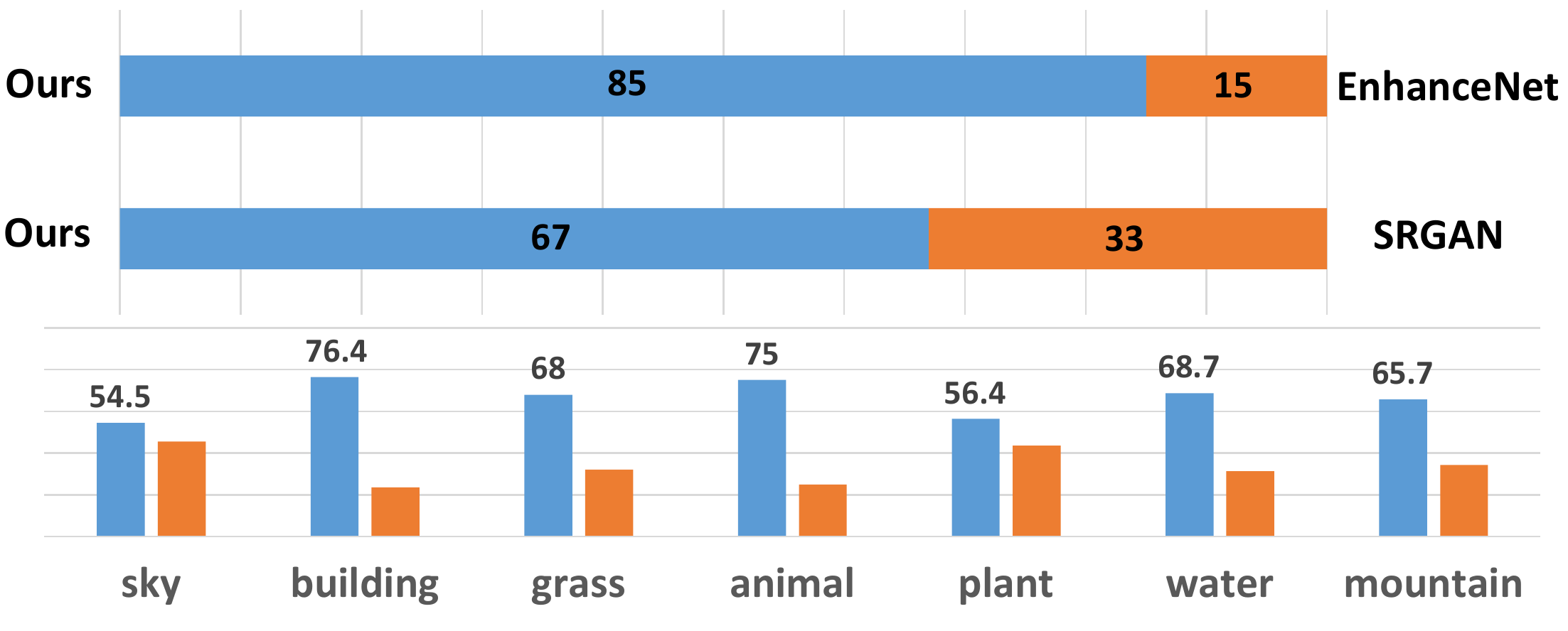}
		\caption{First row: the results of user studies, comparing our method with SRGAN~\cite{ledig2017photo} and EnhanceNet~\cite{sajjadi2017enhancenet}. Second row: our methods produce visual results that are ranked higher in all categories in comparison with SRGAN~\cite{ledig2017photo}.}
		\label{fig:user_study_with_SRGAN}
		\vspace{-0.5cm}
	\end{center}
\end{figure}

We asked 30 users to finish our user study. The results of the first and second sessions are presented in Fig.~\ref{fig:user_study_rank} and Fig.~\ref{fig:user_study_with_SRGAN}, respectively. 
The results of the first session show that SFT-GAN outperforms the PSNR-oriented methods by a large margin.  This is not surprising since PSNR-oriented methods always produce blurry results especially in texture regions. 
Our method sometimes generates good-quality images comparable to HR causing confusion within the users. In the second session, our method is ranked higher than SRGAN~\cite{ledig2017photo} and EnhanceNet~\cite{sajjadi2017enhancenet}, especially in building, animal, and grass categories. Comparable performance is found on sky and plant categories.

\subsection{Ablation Study}
\label{subsec:ablation}

\begin{figure}[t]
	\begin{center}
		\includegraphics[width=1\linewidth]{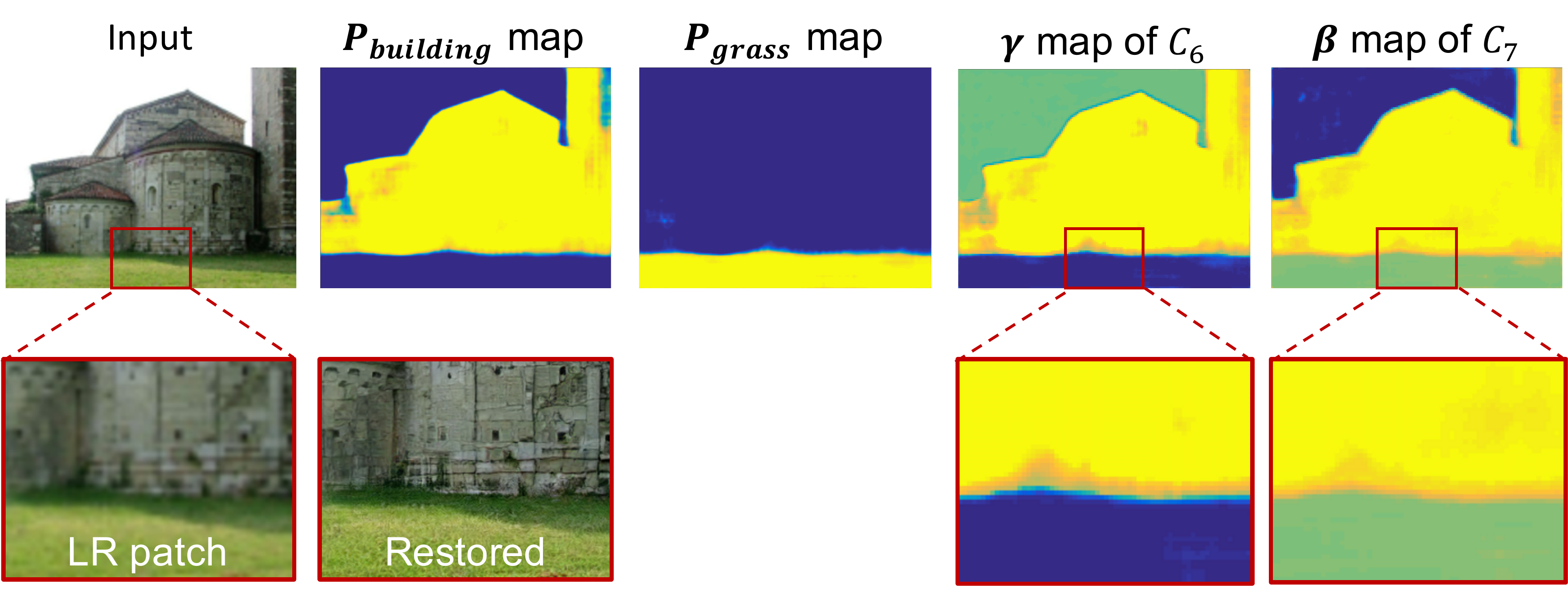}
		\caption{The modulation parameters $\bm{\gamma}$ and $\bm{\beta}$ have a close relationship with probability maps $\bm{P}$ and contain spatial information. The boundary of different categories is clear without interference. $C_i$ denotes the $i^{th}$ channel of the first SFT layer. (\textbf{Zoom in for best view)}}
		\label{fig:probability_to_parameter_map1}
		\vspace{-0.3cm}
	\end{center}
	\begin{center}
		\includegraphics[width=1\linewidth]{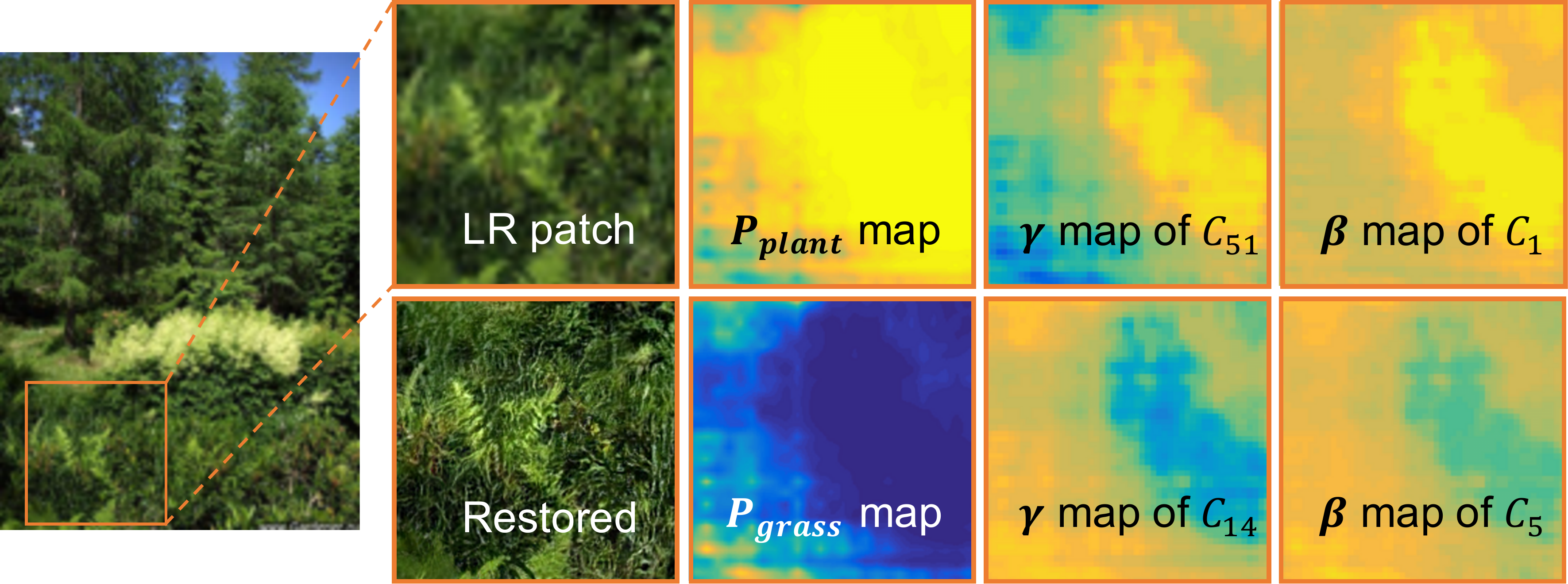}
		\caption{The SFT layer can provide delicate feature transform under the segmentation probability. $C_i$ denotes the $i^{th}$ channel of the first SFT layer. (\textbf{Zoom in for best view)}}
		\label{fig:probability_to_parameter_map2}
		\vspace{-0.8cm}
	\end{center}
\end{figure}

\noindent
\textbf{From segmentation probability maps to feature modulation parameters}. 
Our method modulates intermediate features based on segmentation probability maps. We investigate the relationship between the probability and feature modulation parameters, as depicted in Fig.~\ref{fig:probability_to_parameter_map1}. 
All SFT layers exhibit similar behavior so that we only present its first layer for this analysis.
In the top row, we show an image where building and grass co-exist. 
It is observed that modulation parameters $\bm{\gamma}$ and $\bm{\beta}$ are different for various categorical regions to exert meaningful spatial-wise transformation. 
From the heat map of $\bm{\gamma}$ and $\bm{\beta}$, we can see that the modulation parameter maps are closely related to the segmentation probability maps. 
The boundary of building and grass is clear with a sharp transition. With the guidance of the probability map, our model is able to generate building and grass textures simultaneously without interference of different categories.
Some classes, like plant and grass, may not be clearly distinguishable by their visual appearance. They interlace with each other without a clear boundary.
Despite the ambiguity, the probability maps are still capable of capturing the semantics to certain extent and the SFT layers reflect the subtle differences  between categories in its spatial transformation.
In Fig.~\ref{fig:probability_to_parameter_map2}, the upper row shows the probability map and modulation parameters activated for plant while the lower row shows those for grass. Distinct activations with smooth transition can be observed.
As a result, textures generated by SFT-GAN become more realistic.

\begin{figure}[h]
	\begin{center}
		\includegraphics[width=\linewidth]{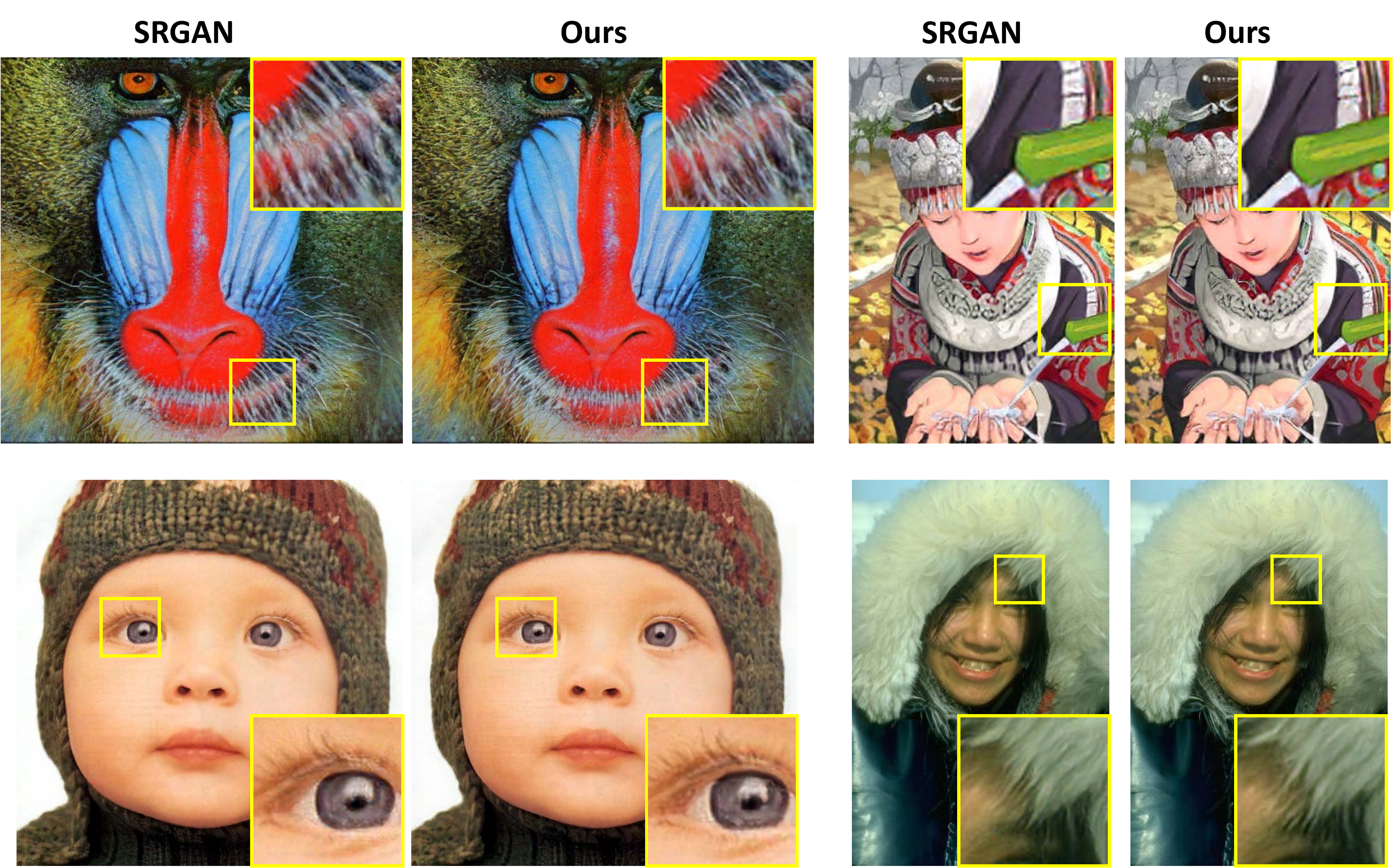}
		\caption{When facing with other scenes or the absence of segmentation probability maps, our model degenerates itself as SRGAN and produces comparative results with SRGAN. (\textbf{Zoom in for best view})}
		\label{fig:degeneration}
		\vspace{-0.4cm}
	\end{center}
\end{figure}

\vspace{0.1cm}
\noindent
\textbf{Robustness to out-of-category examples}. Our model mainly focuses on outdoor scenes and it is effective given segmentation maps of the pre-defined $K$ classes. Despite the assumption, it is also robust to other scenes where segmentation results are not available. 
%
As shown in Fig.~\ref{fig:degeneration}, the SFT-GAN can still produce comparative results with SRGAN when all the regions are deemed as `background'. 
More results are provided in the \textit{supplementary material}.
\vspace{-0.2cm}

\noindent
\textbf{Comparison with other conditioning methods}.
We qualitatively compare with several alternatives for conditioning SR network, which are already discussed in Sec.~\ref{subsec:sfm}.

\noindent
1) \textit{Input concatenation} -- This method concatenates the segmentation probability maps with the LR image as a joint input to the network. This is equivalent to adding SFT conditional bias at the input layer. 

\noindent
2) \textit{Compositional mapping} -- This method is identical to Zhu \etal~\cite{zhu2017your}. It decomposes an LR image based on the predicted semantic classes and processes each region separately using a specific model for that class. Different models share parameters at lower layers.

\noindent
3) \textit{FiLM}~\cite{perez2017film} -- This method predicts one parameter for each feature map without spatial information and then uses these parameters to modulate the feature maps.

\begin{figure}[t]
	\begin{center}
		\includegraphics[width=\linewidth]{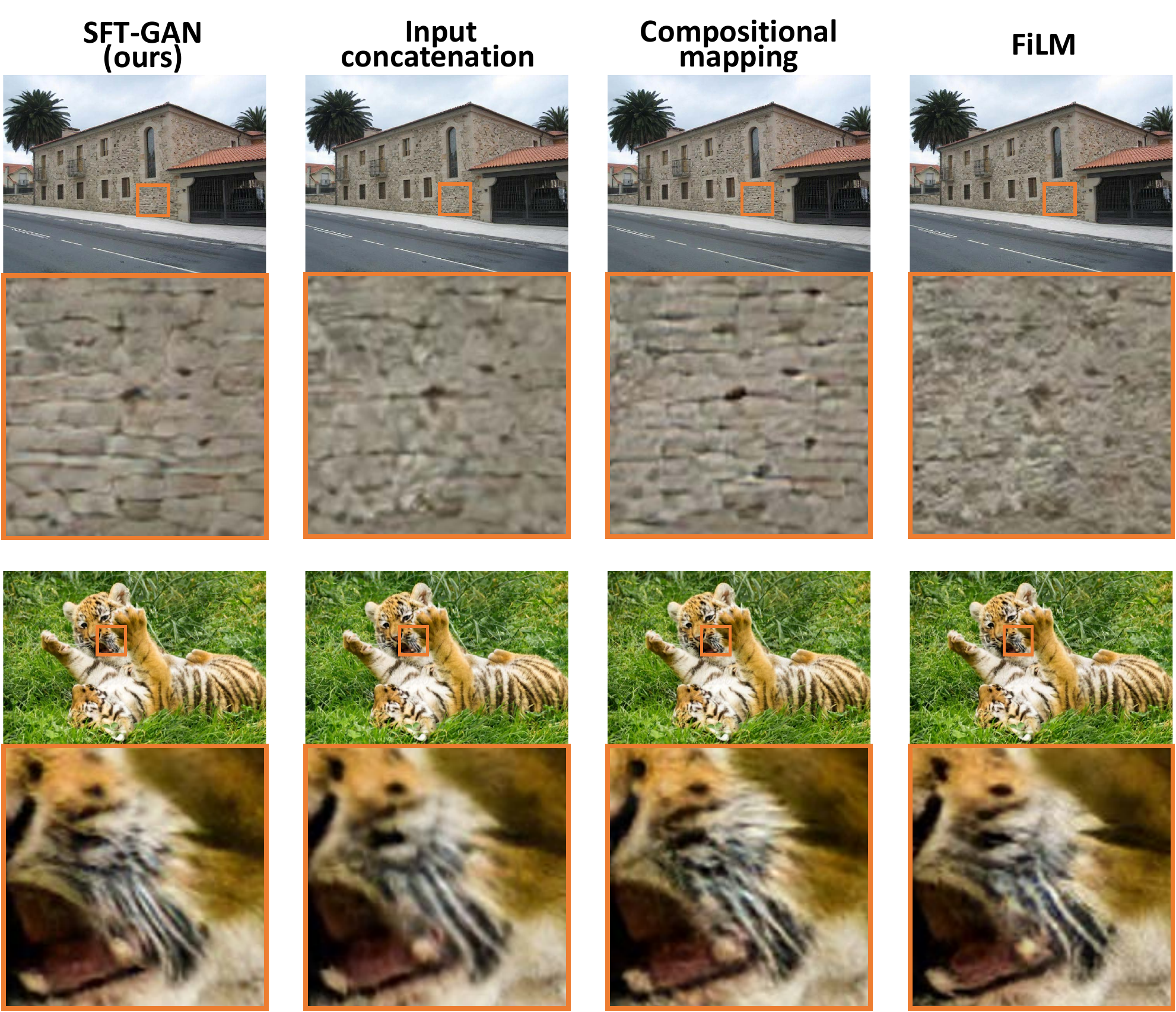}
		\caption{Comparison with other conditioning methods - input concatenation, compositional mapping and FiLM~\cite{perez2017film}. }
		\label{fig:variants}
		\vspace{-0.8cm}
	\end{center}
\end{figure}

As can be observed from Fig.~\ref{fig:variants}, the proposed SFT-GAN yields outputs that are perceptually more convincing. Naive input concatenation is not sufficient to exert the necessary condition for class-specific texture generation. Compositional mapping produces good results but it is not parameter efficient ($\times2.5$ parameters as ours). It is also computationally inefficient as we need to forward several times for a single input image. FiLM~\cite{perez2017film} cannot handle the situations where multiple categorical classes co-exist in an image since it predicts one parameter for each feature map, agnostic to spatial information. For example, in the first image of Fig.~\ref{fig:variants}, the road and sky interfere with the building's structure and thus noisy bricks are generated. Similarly in the second image, the animal's fine texture is severely affected by the grass.

\section{Discussion and Conclusion}
\label{sec:Conclusion}

We have explored the use of semantic segmentation maps as categorical prior for constraining the plausible solution space in SR. A novel Spatial Feature Transform (SFT) layer has been proposed to efficiently incorporate the categorical conditions into a CNN-based SR network. Thanks to the SFT layers, our SFT-GAN is capable of generating distinct and rich textures for multiple semantic regions in a super-resolved image in just a single forward pass. Extensive comparisons and a user study demonstrate the capability of SFT-GAN in generating realistic and visually pleasing textures, outperforming previous GAN-based methods~\cite{ledig2017photo,sajjadi2017enhancenet}.

Our work currently focuses on SR of outdoor scenes. Despite robust to out-of-category images, it does not consider priors of finer categories, especially for indoor scenes, \eg, furniture, appliance and silk. In such a case, it puts forward challenging requirements for segmentation tasks from an LR image. Future work aims at addressing these shortcomings. Furthermore, segmentation and SR may benefit from each other and jointly improve the performance.

\vspace{0.2cm}
\noindent\textbf{Acknowledgement}.
This work is supported by SenseTime Group Limited and the General Research Fund sponsored by the Research Grants Council of the Hong Kong SAR (CUHK 14241716, 14224316. 14209217).

{\small
\bibliographystyle{ieee}
\bibliography{short,bib}
}

\end{document}